# Beyond Optimization: Exploring Novelty Discovery in Autonomous Experiments

Ralph Bulanadi,[1] Jawad Chowdhury,[1] Funakubo Hiroshi,[2] Maxim Ziatdinov,[3] Rama Vasudevan,[1] Arpan Biswas,[4#] Yongtao Liu[1]*

[1] Center for Nanophase Materials Sciences, Oak Ridge National Laboratory, Oak Ridge, TN 37830, USA

[2] Department of Material Science and Engineering, School of Materials and Chemical Technology, Institute of Science Tokyo, Yokohama, 226-8502, Japan

[3] Physical Sciences Division, Pacific Northwest National Laboratory, Richland, Washington, 99352, USA

[4] University of Tennessee-Oak Ridge Innovation Institute, University of Tennessee, Knoxville, TN 37996, USA

* liuy3@ornl.gov

# abiswas5@utk.edu

**Abstract**

Autonomous experiments (AEs) are transforming how scientific research is conducted by integrating artificial intelligence with automated experimental platforms. Current AEs primarily focus on the optimization of a predefined target; while accelerating this goal, such an approach limits the discovery of unexpected or unknown physical phenomena. Here, we introduce a novel framework, INS$^2$ANE (Integrated Novelty Score–Strategic Autonomous Non-Smooth Exploration), to enhance the discovery of novel phenomena in autonomous experimentation. Our method integrates two key components: (1) a novelty scoring system that evaluates the uniqueness of experimental results, and (2) a strategic sampling mechanism that promotes exploration of under-sampled regions even if they appear less promising by conventional criteria. We validate this approach on a pre-acquired dataset with a known ground truth comprising of image–spectral pairs. We further implement the process on autonomous scanning probe microscopy experiments. INS$^2$ANE significantly increases the diversity of explored phenomena in comparison to conventional optimization routines, enhancing the likelihood of discovering previously unobserved phenomena. These results demonstrate the potential for AE to enhance the depth of scientific discovery; in combination with the efficiency provided by AEs, this approach promises to accelerate scientific research by simultaneously navigating complex experimental spaces to uncover new phenomena.

**Introduction**

The recent rise of autonomous experiments (AEs) [1-3], or self-driving laboratories [4-8], is transforming scientific research. By integrating artificial intelligence (AI) and automated laboratory instrumentation, AEs can conduct experiments with minimal human intervention, accelerating materials synthesis [9-11] and characterization processes [12-14]. At the core of AEs are AI approaches, such as Bayesian optimization (BO), which learn continuously from on-the-fly measurements and predict promising experimental parameters that are likely to yield desired properties.

These AI-powered autonomous platforms have demonstrated remarkable success across a wide range of scientific fields. For example, in studies of photovoltaic perovskites, active learning has guided the discovery of perovskite compositions with enhanced performance and stability [15-18]; in thin film deposition, active learning driven autonomous sputtering has shown potential in achieving target composition and enhanced property by close-loop control [19-21]. Beyond material synthesis, self-driving microscopy can pinpoint regions of interest with intriguing physical phenomena via active-learning-driven spectroscopy measurements, reducing time compared to traditional hyperspectral imaging [22-25]; automated microscopy leverages deep learning models to detect predefined objectives from images, driving microscopy to focus on exploration of key regions efficiently [26]. Previous studies have also explored human–AI hybrid strategies to identify curiosity-driven target objectives on-the-fly, relying on real-time human assessment for autonomous materials characterization [27].

However, these current AI approaches in AEs have primarily focused on optimization problems [28,29]. Such models navigate vast parameter spaces and identify materials with superior properties that are predefined by domain experts. These approaches are effective when the experimental objective is very clear, e.g., identifying synthesis conditions or compositions that maximize a known property [30], but fall short when the objective shifts toward more open ended, discovery-oriented tasks. In particular, in autonomous characterization, the goal is not only to optimize but also uncover unexpected or previously unknown physical phenomena that deepen our understanding of the materials [31-32]. Most AEs operate with the assumption that the target is well-defined and known in advance; such AE systems driven by pre-defined targets risk overlooking novel phenomena that fall outside expectations. This behavior limits identification of unprecedented or physically significant phenomena [33-34]. Recently, a subset of the authors proposed a curiosity-driven framework utilizing distinct models and error surrogates for tackling this issue [35]. Here we explore alternative Bayesian approaches with novelty scores that are simple to calculate and implement to create AI frameworks that can not only optimize known objectives but also recognize and prioritize unexpected results [36,37]. Such development could help scientists move beyond incremental improvement and toward knowledge discovery, accelerating the pace at which unknown physics is identified and understood.

In this work, we tackle this challenge and develop an AE approach to maximize novelty discovery. The approach iterates on Bayesian optimization (BO), by integrating a novelty estimation module with a strategic sampling module (Fig. 1). The novelty estimation system analyzes the full experimental dataset in real time and assigns a novelty score to each data point based on its uniqueness relative to other observed results. The strategic sampling mechanism then directs the BO process to explore under-sampled, yet promising regions of the parameter space (rather than oversampling near the global maxima, as in traditional BO), thereby increasing the likelihood of uncovering unexpected phenomena from under-explored regions. We validated this approach using a pre-acquired band excitation piezoresponse spectroscopy (BEPS) dataset from a ferroelectric $PbTiO_3$ sample, where the known ground truth allows for assessment of the effectiveness of the approach. We then implemented the approach within our autonomous scanning microscopy platform, AEcroscopy [38], to explore ferroelectric materials in a real experimental setting. By enabling novelty-driven exploration alongside goal-oriented optimization, this framework advances the capabilities of autonomous experimentation. It represents a step toward scientific discovery systems that are not only faster and more efficient, but also more insightful—capable of revealing the unknown and accelerating our understanding of complex physical systems.

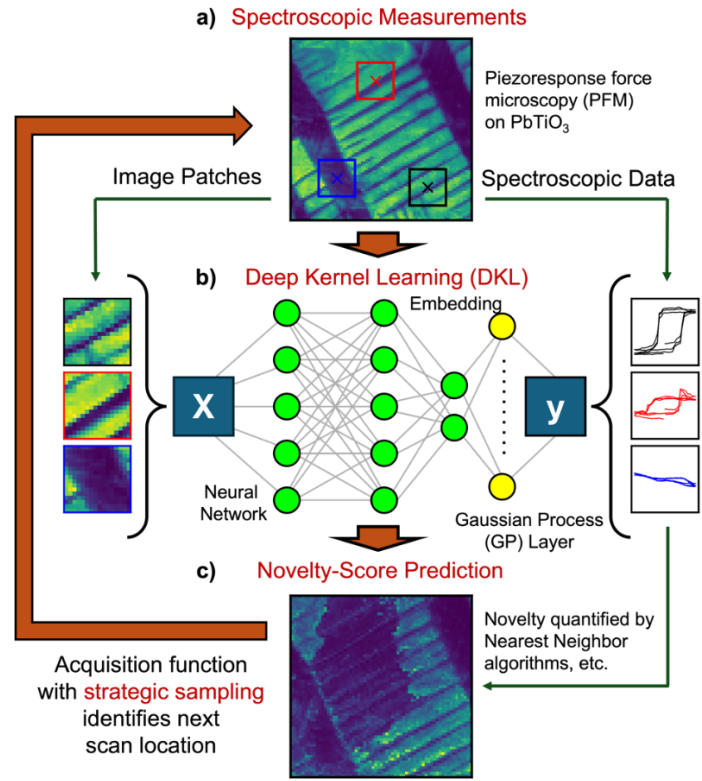

*Figure 1. Novelty discovery in active learning. a) Spectroscopy measurements are taken from a real sample, including both image patches and spectroscopy data. b) Deep-kernel learning is performed to compare image patches to spectroscopy data. c) Novelty-score prediction is performed with strategic sampling to identify new scan regions with the highest expected novelty.*

**Novelty Scoring and Strategic Sampling**

Our general approach for enhanced novelty discovery is shown in Fig. 1. Our experiments begin with the acquisition of a full piezoresponse force microscopy (PFM) image (Fig. 1a) that represents the nanoscale structure. We then acquire spectroscopic data (i.e., hysteresis loops) at a few selected locations, then extract predefined scalar descriptors from this spectroscopic data. We then employ deep kernel learning (DKL) [23, 39], which combines the representational power of neural networks with Gaussian process uncertainty quantification, to analyze the relationship between local ferroelectric domain structure (from PFM images) and polarization–switching characteristics (from BEPS hysteresis loops) (Fig. 1b). An acquisition function determines the next spectroscopy measurement location, balancing exploration and exploitation based on the DKL prediction and uncertainty (Fig. 1c).

In this classical setup, the AE is guided by a pre-defined scalar descriptor based on existing physics knowledge. For example, given the PFM structure image in Fig. 2a, hysteresis loops can be gathered at particular points (Fig. 2b). A physical descriptor such as loop area (Fig. 2c) can then be used as the scalarizer for model training and optimization in the acquisition function. While this method is effective in optimizing measurements when the system under investigation matches expectation [40], it may overlook unprecedented features that fall outside of expectation, e.g., if new features appear in hysteresis loops that are unrelated to nucleation bias or coercive bias.

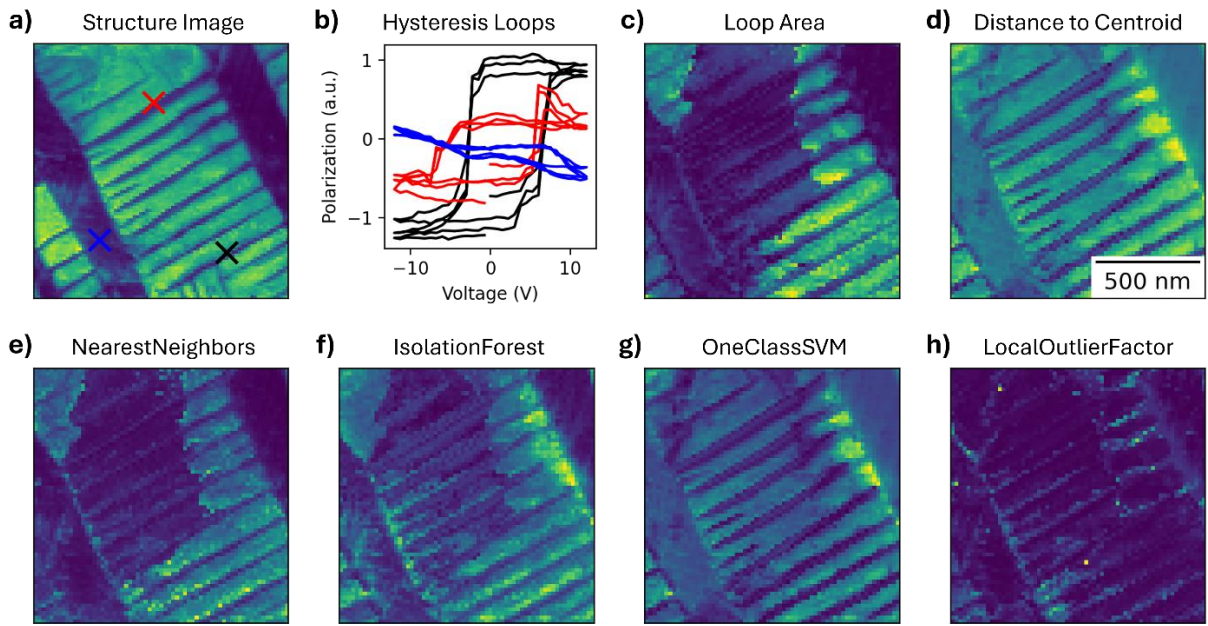

*Figure 2. Scoring and acquisition approaches applied on a) The original structure images. Crosses represent regions where b) sample hysteresis loops are taken to generate c) the loop area, which is the used physical scalarizer. The novelty scores used are d) Distance to Centroid; e) NearestNeighbors; f) IsolationForest; g) OneClassSVM; and h) LocalOutlierFactor. In a,c-h) darker colors are lower (arbitrary) values; lighter colors are higher (arbitrary) values.*

Therefore, instead of searching for the physical descriptor, we can instead integrate a scoring system that evaluates the novelty of spectroscopic data to enable the identification of unforeseen phenomena in real time AEs. Given the diversity of material systems that could be analyzed, we implemented five distinct approaches to novelty assessment: Distance to Centroid (DtC), Nearest Neighbors (NN), IsolationForest (IF), OneClassSVM (OC-SVM), and LocalOutlierFactor (LOF). The later four methods use implementations from scikit-learn [41]. Each of these approaches assess different local and global conditions in the dataset to create their novelty scores. The details of each method can be found in the Supplementary material.

These five novelty assessment approaches were first applied to the entirety of a pre-acquired BEPS dataset that has previously been used for machine learning (ML) development. This dataset contains 10000 BEPS hysteresis loops which map different domain structures, including domains that are oriented either out-of-plane or in-plane with respect to the sample surface, as well as various domain walls. The out-of-plane domains can be identified as regions with a higher BEPS signal, while the in-plane regions are those with a lower BEPS signal; the domain-walls are the boundaries between adjacent domains. Supp. Fig. 1 is an annotated image of these structures.

The DtC novelty score (Fig. 2d), in prioritizing spectroscopy data that is globally distinct from others, appears to most highly prioritize the out-of-plane domains, likely due to their state as the most extreme, highest-signal regions. The in-plane domains are also given intermediate priority, while the domain walls, which compose the smallest minority of the dataset, appear to be considered the least novel. In the NN novelty score map (Fig. 2e), one of the out-of-plane domains expresses high novelty (while the other expresses lower novelty), and the in-plane domains express the lowest novelty. Most interestingly, domain walls tend to express higher novelty. The IF novelty score map (Fig. 2f) is similar to NN, but its focus on domain walls is yet more prominent. While IF continues to prioritize one out-of-plane domain over the other, the difference is also less stark here. IF, in prioritizing extrema of the entire dataset, appears to focus on hysteresis loops that have at least some unique behaviors, even if most of the behavior of that same hysteresis loop is typical. The OC-SVM (Fig. 2g) novelty score map is similar to the DtC novelty score map, likely because they are both largely global assessments of the dataset, without the focus on extrema. The LoF novelty score (Fig. 2h) is rather unique, in that the domain walls alone are considered novel, while the domains themselves are considered less so.

While all of these novelty scoring systems may have some use in particular experiments, proper novelty scoring methods can be selected based on domain knowledge. We recognize that our in-plane domains, in presenting a weaker BEPS signal, have a lower signal-to-noise ratio (SNR) than out-of-plane domains. Since we want to measure novel data but not necessarily noisy data (which may also appear unique), an effective novelty detection method should avoid misclassifying low signal-to-noise ratio (SNR) data as novel. As a result, we exclude the DtC and OC-SVM approaches, as they assign mid-level novelty scores to these in-plane domain regions, indicating a failure to properly account for the low SNR. While LoC could be useful on a study particularly focused on domain-wall assessment, it is not a good choice for more exploratory studies in which domains would also be investigated. Thus, we consider both NN and IF novelty scores to be more suitable for autonomous exploration of this dataset—with the caveat that while the remaining novelty score assessments are not appropriate here, they could be potentially good choices for studies in other physical systems.

While using novelty scoring as a metric has key potential to improve scientific discovery, traditional acquisition functions in BO still have key issues in that their goal is still inherently optimization—this risks a process continually searching the same "novel" region of parameter space. For the purposes of the discovery and exploration of novel phenomena, the search for a single promising optimum region may be counterproductive to true exploration of the entire sample space. To overcome this limitation, we employed a strategic sampling approach, SANE, to promote exploration of under-sampled locations even if they are neither most promising nor uncertain. SANE accomplishes this by applying a non-uniform cost function that promotes searching new regions of interest; more details about SANE can be found in our previous work [42]. An Integrated Novelty Score–SANE (INS$^2$ANE) approach has the potential for improved diversity in exploration and discovery of unknown and novel phenomena.

**Implementation with a Pre-Acquired Model Dataset**
With the suitability of various novelty-scoring metrics assessed over the entire model dataset, we could then assess the effectiveness of novelty-score guided processes against a traditional physical-descriptor-driven approach. We thus carried out three AEs on the model dataset: (1) A standard AE driven to optimize the physical descriptor; (2) a standard AE driven by novelty scores, and (3) an INS$^2$ANE AE that performs strategic sampling based on novelty scores. We used both IF and NN novelty scores as a metric to assess novelty with respect to real-time acquired spectroscopic data.

The results of a standard AE, with the area of the ferroelectric hysteresis loop as the physical descriptor, are shown in Fig 3(a–b). As shown in Red–White in Fig. 3b, the system quickly begins scanning in out-of-plane domains, near walls with an in-plane domain, then rarely performs exploratory measurements in other regions. In comparison with the ground truth in Fig. 3a, it is apparent that the AE finds many of the regions with the largest loops. However, the predicted scalarizer, shown in Blue–Green in Fig. 3b, is particularly distinct from the ground truth. The AE predicts large responses in all out-of-plane regions near domain walls, without distinction between up and down domains; while the ground truth shows that the out-of-plane domain towards the bottom of the image has a higher response than the out-of-plane domain in the center of the image.

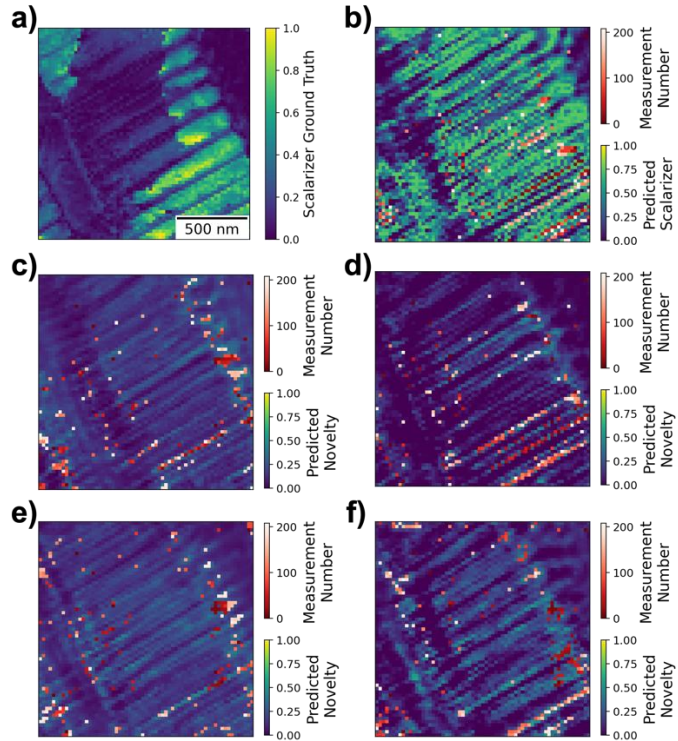

*Figure 3. Implementation of novelty-driven AEs on a pre-acquired dataset. a) shows the ground truth for the physical scalarizer shown in b); c) uses IF-novelty scores; d) uses NN-novelty scores; e) uses IF-INS$^2$ANE; and f) uses NN-INS$^2$ANE. For b–f) the predicted scalarizer/novelty is shown in Blue–Green after the final measurement, with the number in sequence at which the point was scanned is shown in Red–White.*

For novelty-driven AEs the system more frequently and more continually measures diverse regions and domains. IF novelty score, a global metric, is used in Fig. 3c, while NN novelty score, a local metric, is used in Fig. 3d. In both cases, unlike the scalarizer-driven AE, the novelty-driven AE continually switches between different regions, with particularly more time spent examining large in-plane domains, and out-of-plane regions near the in-plane domains. The main apparent difference between IF-novelty-driven and NN-novelty-driven AEs are that the global IF novelty score drives the AE to measure regions of interest more densely, resulting in larger regions in which no measurements are taken at all; in contrast, the local NN novelty score drives the AE to measure more sparsely, yielding a more general view of the system.

Finally, the INS$^2$ANE AE, presented in Fig. 3(e–f), shows distinct but situationally advantageous behavior. SANE attempts first to classify local maxima, before moving and finding other distinct local maxima. Priority is thus given to nearby regions of interest, before intermittently (here, every 5$^{th}$ measurement) shifting to investigate more remote regions of interest to search for a new local maxima. INS$^2$ANE therefore measures points densely around a particular (novel) region of interest—and therefore provides a more holistic understanding of the locality—before moving to a new region of interest. In practice, this results in stratification of the behavior described for novelty-driven AEs, where both IF (Fig. 3e) and NN (Fig. 3f) show dense sampling around particular regions of interest, and sparse sampling throughout other regions in the dataset.

**Quantitative Assessment of Novelty-Driven AEs**

To quantitatively evaluate this INS$^2$ANE approach to novelty discovery, we employed two complementary methods: first, we measured how accurately each AE approach predicted physical scalarizers in the system (Fig. 4a), which we use as an abstraction for physical understanding; and second, we analyze the variability of the measured hysteresis loops (Fig. 4b) to ensure our AE is truly examining distinct and anomalous regions of the sample space.

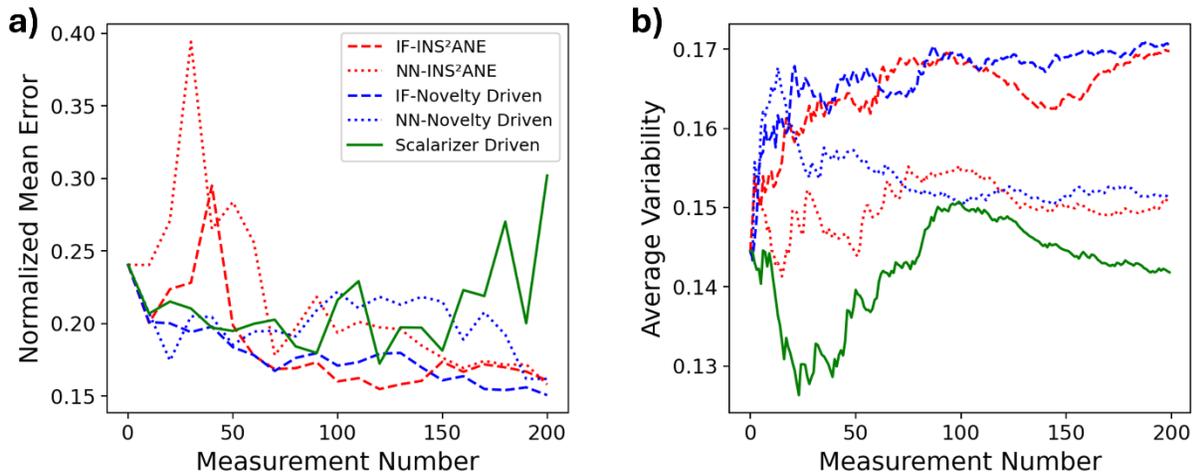

Figure 4. Assessment of the novelty-driven AEs. a) Comparison based on NME of the physical scalarizer prediction. b) Comparison based on variability of measured hysteresis loops.

Fig. 4a shows the variation in normalized mean error (NME) for each AE approach across 200 iterations. The initial NME of all our systems after the initial 10 random samples are taken is 0.24. After this, our scalarizer-driven system has an initial drop in the NME, decreasing to a minimum of 0.18 after 80 additional samples. However, while this system seems to have a strong understanding of high-scalarizer regions, the overall understanding of the system is lacking and rather appears to be overfitted. This is evident in the increase of NME, until it reaches a maximum of 0.30 in the final iteration. In contrast, the IF-novelty-score-driven AE generally drops and reaches a minimum of 0.15, while the NN-novelty-score-driven AE reaches 0.16. Both novelty-driven AEs therefore show substantial improvement in physical understanding of the system, and appear to drop through the end of the 200 measurements. The INS$^2$ANE approaches show rather different behavior, however: the NME rather is increased, with IF-novelty scoring reaching a maximum of 0.29 after 40 measurements, and NN-novelty scoring reaching a maximum of 0.39 after 30 measurements, likely due to the system overfitting physics understanding specifically around the anomalous points (as intended). Both systems then rapidly reduce their NME, with both reaching a minimum of 0.16 after 200 measurements. IF-INS$^2$ANE presents the lowest NME of all approaches between 80 and 140 measurements, but all novelty-driven methods end with significantly lower NMEs than the scalarizer-driven AE. These results suggest that novelty-driven approaches promote more thorough and comprehensive investigation of the system by capturing a wider diversity of data, which in turn enhances model prediction accuracy.

The variability of the hysteresis loop datasets is shown in Fig. 4b. We defined variability as the mean of the standard deviation of the measurement at each timestep among all hysteresis loops. Larger variability indicates that each measured hysteresis loop is more unique with respect to other measured loops. The initial 10 measurements used in all AEs had a variability of 0.14. The scalarizer-driven AE had an initial reduction in variability—the system searches continually for a particular optimal structure, which results in similar structures and therefore a lower variability. However, once all relevant regions are found, the variability of the dataset increases and begins to approach the starting variability of 0.14. With novelty score as the optimization parameter, the variability immediately spikes to a maximum. The NN-novelty-driven AE then reduces its variability, while the IF-novelty-driven AE continues to rise through the extent of the experiment; these reach a variability of 0.15 and 0.17 respectively. INS$^2$ANE, yields a similar, but smoother rise in variability than the respective classic AEs but regardless approaches a similar final variability (NN-INS$^2$ANE: 0.15; IF-INS$^2$ANE: 0.17) to their equivalents that do not employ strategic sampling.

These measured variabilities are particularly notable when compared to the ground truth baseline. The variability of the entire model dataset was calculated to be 0.141. From 2000 realizations, the average variability calculated from 10 initial random points is initially lower, at approximately 0.131 (standard deviation: 0.015), but reaches an average variability of 1.404 (standard deviation: 0.003) after sampling 200 random points (Supp. Fig. 2). Here, all novelty-driven AE have a variability substantially higher than would be obtained from random selection, while the scalarizer-driven AE falls within random expectation.

**Implementation in Autonomous Microscopy**
With these approaches evaluated in our pre-acquired dataset, we then directly implemented these approaches into an autonomous scanning probe microscopy platform using the AEcroscopy ecosystem [38]. As with the pre-acquired dataset, we investigated a ferroelectric thin film with diverse polarization configurations using piezoresponse spectroscopy measurements. 10 points were randomly sampled, and the BO was then used after every step to direct new tip motion and data acquisition, as with our implementation on our pre-acquired model. This proceeded until 100 hysteresis loops were acquired. Realizations were performed using a physical-descriptor-driven AE, a novelty-score-driven AE, and an INS$^2$ANE AE. Unlike with pre-acquired data, each AE process cannot be performed on identical datasets; as such, four realizations were performed for each process at different locations to avoid the effects of region-to-region variation in measurements—1 realization was performed with an IF novelty score (Fig. 5), and 3 realizations were performed with an NN-novelty score (Supp. Fig. 3).

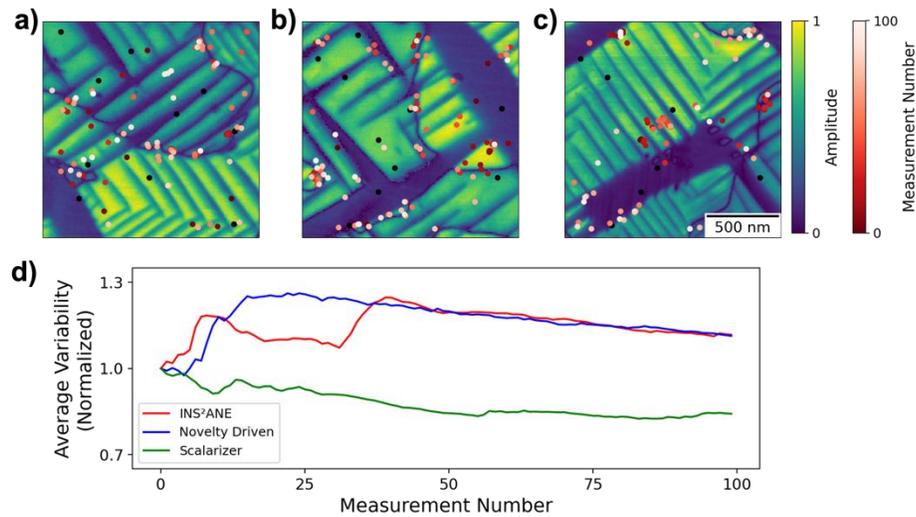

*Figure 5. Novelty-driven AEs as run on a real scanning probe microscope. a) shows a standard scalarizer-driven AE; b) shows an IF-novelty-driven AE; and c) shows the IF-INS$^2$ANE AE. Images show the piezoresponse amplitude as the structure image. Dots represent locations measured and their order in sequence; dots in black were randomly selected for training. d) shows the variability in collected hysteresis loops among all the different AEs.*

As with the pre-acquired model dataset, the classic AE with physical descriptors quickly finds and acquires data at domain walls (Fig. 5a): where the easiest, largest-area hysteresis loops are expected. Virtually no data is taken at the in-plane domain in the lower right, and the out-of-plane domain towards the bottom is relatively unexplored. The novelty-driven AE (Fig. 5b) also applies some focus to the out-of-plane domain wall, but relatively more time is spent around the large in-plane domain as well. The INS$^2$ANE AE characteristically measures rather distinct locations more densely (Fig. 5c), with one central scan "cluster" measured early on in the center, and later scan clusters around each type of domain wall.

The normalized variability of each AE type is shown in Fig. 5d. The scalarizer-driven AE shows a gradual drop in variability, akin to what was observed in the latter half of the execution on the pre-acquired dataset. Both the novelty-driven AE and the INS$^2$ANE AE in contrast show a sudden jump in variability, before a slow decay. Notably, INS$^2$ANE shows a second jump in variability around step-30, which can be attributed to the discovery of another novel region due to strategic sampling. This shows that novelty scoring, with or without strategic sampling, allows for more diversity in measured data, while strategic sampling has the potential to push the boundary for further novelty discovery.

This integrated framework positions autonomous scanning probe microscopy as not only an efficient nanoscale characterization tool but also a strategic, intelligent platform for scientific discovery, capable of identifying emergent behaviors and revealing previously unobserved nanoscale phenomena.

## Conclusions and Future Directions

In summary, we developed an approach, INS$^2$ANE, to enable novelty discovery in autonomous experimentation, aimed at facilitating the identification of new physical phenomena through the detection of novel results. The approach integrates two key components: a novelty scoring system and a strategic sampling system. We validated this method using both a model dataset and real-world autonomous experiments on a ferroelectric thin film sample. Our results demonstrate that both systems independently contribute to the discovery of novel data, though their effectiveness varies across different stages of the autonomous experimentation process. The novelty scoring system operates on the result space, identifying and prioritizing previously unseen or unexpected outcomes. In contrast, the strategic sampling system targets the parameter space, encouraging exploration of regions that are under-sampled or poorly characterized. By integrating this approach to novelty discovery with a curiosity-driven framework previously developed by some of the authors to enhance structure-spectroscopy analysis, this work has the potential to shift AE from a focus on optimization to a pathway for uncovering fundamental understanding [35].

Finally, we observe that when one system reaches a point of diminishing returns in terms of novelty discovery, the other often becomes more effective, pushing the boundary of novelty further. However, the timing of when each system performs better cannot be determined in advance. This observation suggests a promising direction for future research is the integration of human-AI collaboration to dynamically switch between the two systems, as well as tuning novelty assessment metrics and integrating human intuition, during autonomous experimentation. Such adaptive control and meta-learning strategies could significantly enhance the efficiency and depth of scientific discovery in complex experimental spaces.

## Acknowledgements

This research and workflow development was sponsored by the INTERSECT Initiative as part of the Laboratory Directed Research and Development Program of Oak Ridge National Laboratory, managed by UT-Battelle, LLC for the US Department of Energy under contract DE-AC05-00OR22725. Piezoresponse force microscopy was performed at and supported by the Center for Nanophase Materials Sciences (CNMS), which is a US Department of Energy, Office of Science User Facility at Oak Ridge National Laboratory. Development of the SANE algorithm acknowledges the use of facilities and instrumentation at the UT Knoxville Institute for Advanced Materials and Manufacturing (IAMM) and the Shull Wollan Center (SWC), supported in part by the National Science Foundation Materials Research Science and Engineering Center program through the UT Knoxville Center for Advanced Materials and Manufacturing (DMR-2309083). The authors acknowledge Ganesh Narasimha for valuable discussions and insightful suggestions.

## Conflicts of Interest

The authors declare no conflicts of interest.

**Authors Contribution**

Y.L. conceived the idea. R.B. and Y.L. conducted the investigation. M.Z. developed DKL. A.B. developed the SANE framework. F.H. synthesized the $PbTiO_3$ sample. R.B. and Y.L. wrote the manuscript. All authors edited the manuscript.

**Data Availability**

The code and the notebooks used in this work can be found on GitHub; see Ref. [43].